\pdfoutput=1

\documentclass[11pt]{article}

\usepackage[]{acl}

\usepackage{times}
\usepackage{latexsym}

\usepackage{graphicx}
\usepackage[T1]{fontenc}

\usepackage[utf8]{inputenc}

\usepackage{microtype}

\usepackage{inconsolata}

\usepackage{booktabs}

\title{A Survey of Data Synthesis Approaches}

\author{Hsin-Yu Chang$^*$ \quad Pei-Yu Chen$^*$\quad Tun-Hsiang Chou$^*$\quad Chang-Sheng Kao$^*$\\
\textbf{Hsuan-Yun Yu$^*$\quad Yen-Ting Lin\quad Yun-Nung Chen}\\
        National Taiwan University, Taipei, Taiwan \\ 
        \texttt{\{r12944014, r12922045, r11922163,r11922a14, r12922121\}@ntu.edu.tw} \\
        \texttt{\{ytl, y.v.chen\}@ieee.org}}

\begin{document}
\maketitle
\begin{abstract}

This paper provides a detailed survey of synthetic data techniques. We first discuss the expected goals of using synthetic data in data augmentation, which can be divided into four parts: 1) \textit{Improving Diversity}, 2) \textit{Data Balancing}, 3) \textit{Addressing Domain Shift}, and 4) \textit{Resolving Edge Cases}. Synthesizing  data are closely related to the prevailing machine learning techniques at the time, therefore, we summarize the domain of synthetic data techniques into four categories: 1) \textit{Expert-knowledge}, 2) \textit{Direct Training}, 3) \textit{Pre-train then Fine-tune}, and 4) \textit{Foundation Models without Fine-tuning}. Next, we categorize the goals of synthetic data filtering into four types for discussion: 1) \textit{Basic Quality}, 2) \textit{Label Consistency}, and 3) \textit{Data Distribution}. In section 5 of this paper, we also discuss the future directions of synthetic data and state three direction that we believe is important: 1) focus more on quality, 2) the evaluation of synthetic data, and 3) multi-model data augmentation.\footnote{\url{https://github.com/MiuLab/SynData-Survey}}
\begingroup\def\thefootnote{\rm *}\footnotetext{Equal contribution.}\endgroup

\end{abstract}

\section{Introduction}
\begin{figure*}[!ht]
    \centering
    \includegraphics[width=1\linewidth]{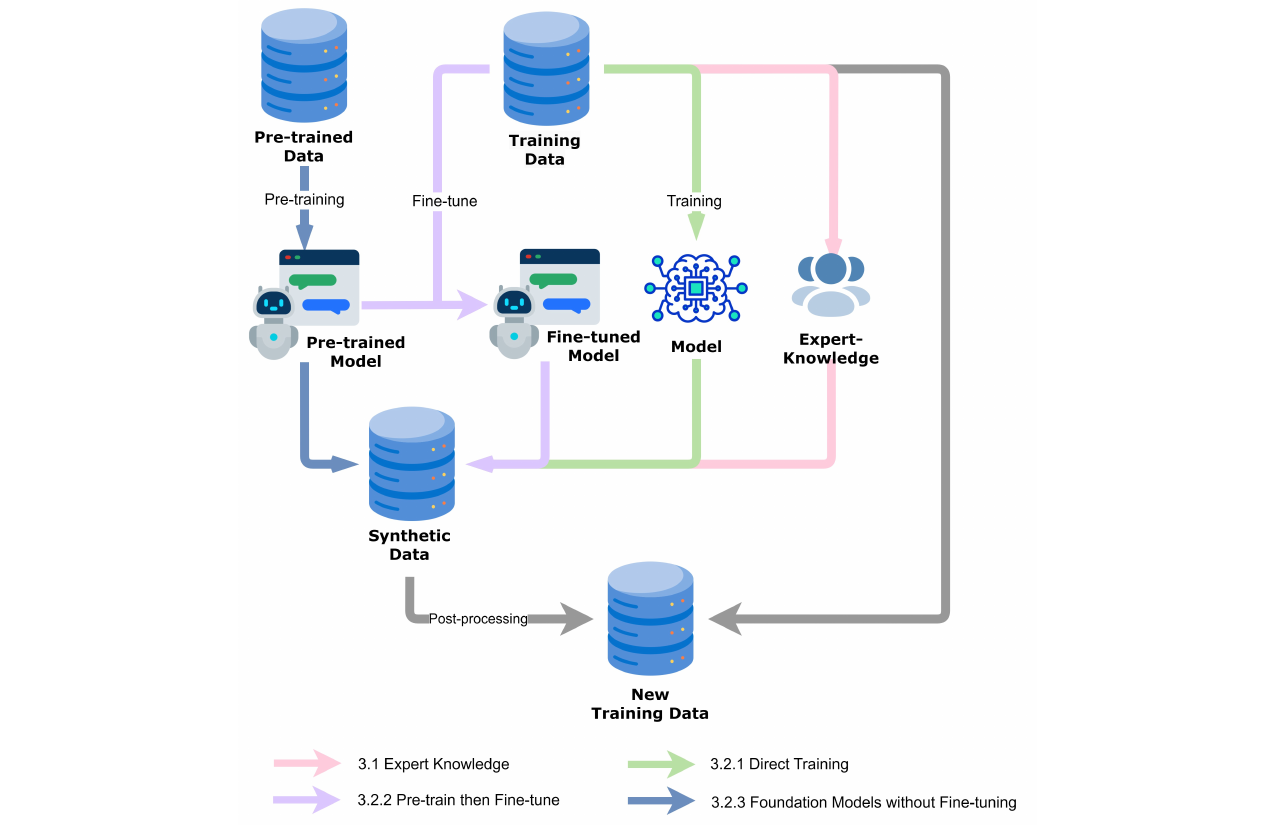}
    \caption{Four approaches to generate synthetic data: 1) Expert-Knowledge, 2) Direct Training, 3) Pre-train then Finetune, and 4) Foundation Models without Fine-tuning. Each approaches are discussed detailedly in section 3.}
    \label{fig:approaches2}
\end{figure*}


Synthetic data has always played a significant role in the field of machine learning \cite{he2008adasyn, bolon2013review}.
With the development of machine learning, the techniques used for generating synthetic data are also advancing rapidly. In general, we can divide the pipeline for obtaining synthetic dataset into two stages: \textit{Synthetic Data Generation} and \textit{Post-processing}, as shown in Figure \ref{fig:approaches2}. In the stage of \textit{Synthetic Data Generation}, it is mainly achieved through methods such as modifying existing data, annotating unlabeled data, or directly generating new data. During the \textit{Post-processing} stage, the main objective is to filter out inappropriate data to ensure that synthetic data can be beneficial for subsequent data augmentation processes.

In this survey paper, we aim to re-explore the following points from the different perspectives: 1) the objectives of data augmentation, 2) the approaches to synthetic data generation, and 3) the benefits of synthetic data filtering. We first explore augmentation objectives, which are the reasons behind conducting data augmentation and what problems it aims to solve.
We categorize these objectives into four types: \textit{Improving Diversity}, \textit{Data Balancing}, \textit{Addressing Domain Shift}, and \textit{Resolving Edge Cases}, as discussed in Section \ref{section: augmentation objectives}.
Next, we intend to explore different approaches to synthetic data generation by categorizing them based on the technological advancements in different periods: starting from directly training a model, to training a foundation model and then fine-tuning it, and finally using a foundation model directly.
We found a high correlation between the techniques used for obtaining synthetic data and the machine learning techniques that were popular during the same period.
Therefore, similar to  \citet{prompt-tuning-survey}, we divided the eras of synthetic data techniques into four periods: \textit{Expert Knowledge}, \textit{Direct Training}, \textit{Pre-train then Fine-tune}, and \textit{Foundation Model without Fine-tuning}. For a more detailed explanation, please refer to Section \ref{section:approaches}. As for Section \ref{section:filtering}, we categorize synthetic data post-processing into three types based on their purposes: \textit{Basic Quality}, \textit{Label Consistency}, and \textit{Data Distribution}. In the Synthetic Data Post-processing section~\ref{section:filtering}, our focus lies primarily on the considerations required to filter out the data obtained during the synthetic data generation stage, ensuring that the entire synthetic dataset is beneficial for aiding data augmentation. In the past, there have been many outstanding survey papers on synthetic data\cite{deepmind-synthetic-survey, llm-era-synthetic-survey, synthetic-survey-2021, synthetic-survey-what}, and our work builds upon these foundations to offer a new perspective on this evolving field.

\section{Augmentation Objectives}\label{section: augmentation objectives}

In this section, we categorize these objectives into four types: \textbf{Improving Diversity}, \textbf{Data Balancing}, \textbf{Addressing Domain Shift}, or \textbf{Resolving Edge Cases}. A single data augmentation method may not be limited to addressing only one of the objectives mentioned above.

\subsection{Improve Diversity}

Previous works have found that simply increasing the training data size through data augmentation can often lead to overfitting during subsequent model training. Therefore, enhancing data diversity can make it more difficult for the models to fit to the augmented data, resulting in better generalization capabilities \cite{affinity-and-diversity}. \citet{randaugment} employs random sampling of transformation subsets to reduce the search space for data augmentation methods while maintaining the diversity of augmented data, thereby enhancing model performance. \citet{controllable-and-diverse-sum} addresses the issue of insufficient diversity in generated dialogues due to a lack of seed dialogues by leveraging the LLMs' in-context learning capability. They generate diverse dialogue summaries based on this and then use them as a foundation to generate rich and diverse open-domain dialogues. \citet{wang-etal-2023-self-instruct} utilizes Rouge-L \cite{lin-2004-rouge} to examine the similarity between generated data. The study employs filtering to remove excessively similar generated data, thus maintaining the diversity of the synthetic dataset. Additionally, it addresses the limitation of manually acquired instruction-following data, which tends to be restricted to specific tasks.

\subsection{Data Balancing}
Data imbalance in machine learning leads to models that are biased towards the majority class, often resulting in poor generalization for minority classes. Techniques like resampling or synthetic data generation can mitigate these effects by providing more balanced training data.  Pioneering techniques like SMOTE (\citealp{Chawla_2002}) or ADASYN (\citealp{adasyn}) generate synthetic examples for minority classes rather than just replicating existing ones, providing more balanced examples for the model to learn from.

\subsection{Address Domain Shift}
Domain shift occurs when a model trained on a source task needs to make predictions on a different target domain. Data augmentation adapts the model by expanding the source dataset to better generalize across domains. This involves generating additional training samples, like rotated or deformed images with added noise, to increase robustness to variations.

\citet{chen2021data} enhances model learning by injecting noise (such as random shuffling, loss, or masking of some words) into input sentences, training the model to recover information from the perturbed inputs. Besides, the study also transforms sentences from the source domain into the target domain's format and then reconstructing them back to their original source domain form, teaching the model how to map and transform data between different domains. \citet{orbesarteaga2022augmentation} improves model performance on data from different domains using consistency training and adversarial learning. The former involves randomly augmenting input images and requiring the model to produce consistent predictions for the augmented images. The latter uses adversarial networks to differentiate features from different domains, thereby prompting the model to generate features that cannot be distinguished by the adversarial networks.

\subsection{Resolve Edge Cases}
Addressing edge cases, scenarios that occur at the extreme ends of data distributions, is crucial for achieving comprehensive and reliable performance \cite{2019ISPAn42W7}. Data augmentation serving as a strategic approach to synthetically expand the variety of training data by introducing rare but plausible scenarios. This technique not only enriches the dataset but also ensures that the model is exposed to and learns from these edge cases, thereby reducing the likelihood of erratic behavior or misclassifications in less common conditions. By simulating various real-world perturbations and anomalies, data augmentation effectively enhances the model's generalizability and resilience, making it adept at handling a wider array of situations, which is particularly beneficial in critical applications such as medical imaging, autonomous driving, and anomaly detection in cybersecurity. \cite{yudkin2022handsup}

\section{Augmentation Approaches}\label{section:approaches}
In this section, we introduce various approaches to generate synthetic data, and categorize these approaches into four types: \textbf{Expert Knowledge}, \textbf{Direct Training}, \textbf{Pre-train then Fine-tune}, and \textbf{Foundation Models without Fine-tuning}. Four different approaches are shown in table \ref{tab:approaches}.




\renewcommand{\arraystretch}{1.5}
\begin{table*}
    \small
    \centering
    \begin{tabular}{p{3cm}p{4.5cm}p{7.5cm}}
        \toprule
        \bf Approach & \bf Concept & \bf Examples \\
        \midrule
        Expert Knowledge & Create new examples with human knowledge. & \citet{zhang2016characterlevel}, \citet{eda}, \citet{wiki}, \citet{aeda} \\
        \hline
        Direct Training & Train a model on data specific to the task for synthesizing new data. & \citet{Fadaee_2017}, \citet{kobayashi2018contextual}, \citet{xu2016improved}, \citet{guo2019augmenting} \\
        \hline
        Pre-train then Fine-tune & Pre-train a model on a large dataset and then fine-tune the pre-trained model to the target task to create new data.  & \citet{xu2021augnlg}, \citet{doubinsky2023semantic}, \citet{samuel2023generating}, \citet{chen2024generalizable}, \citet{kumar-etal-2020-data} \\
        \hline
        Foundation Models without Fine-tuning & Augment new data with foundation models directly without fine-tuning. & \citet{auggpt}, \citet{abdullin2024synthetic}, \citet{z-icl}, \citet{liu-etal-2022-wanli}, \citet{sahu-etal-2022-data}, \citet{lee-etal-2022-personachatgen}, \citet{wang-etal-2023-self-instruct}, \citet{honovich-etal-2023-unnatural} \\
        \bottomrule
    \end{tabular}
    \caption{Different approaches and related examples}
    \label{tab:approaches}
\end{table*}

\subsection{Expert Knowledge}
Expert knowledge based engineering is a traditional technique within machine learning aimed at selecting, extracting, and transforming raw data into new features based on the domain expertise of researchers or engineers \cite{Guyon2006feature}. The primary objective of expert knowledge-based engineering is to augment the performance of models by furnishing more relevant features despite constraints in data availability. This process entails the creation, transformation, extraction, and selection of features—also referred to as variables—that are best suited to optimize the accuracy of machine learning algorithms.


\begin{enumerate}
    \item Creation:
    The methods include synonym replacement \cite{eda, zhang2016characterlevel}, where certain words in the text are replaced with their synonyms to create a new text representation. Another approach is the random insertion of words or phrases \cite{eda}. Both Approaches can help the model better understand different expressions and thus improve its generalization ability. There are also methods that use existing data to set rules \cite{wiki} and transform it into the desired training data format, such as converting the established format of Wikipedia into a dataset required for QA.

    \item Transformation:
    The methods include dispersing punctuation marks throughout the text \cite{aeda}, changing the structure or format of the original text. This approach retains the essence of the content while still presenting different sentence structures to the learning algorithm.
    
    \item Hybrid:
    Mapping the data from a specific domain to the distribution of a general domain, and conducting data augmentation by finding similar data in the general domain, is equivalent to combining feature transformation and feature creation. This approach is known as domain adaptation or transfer learning in machine learning. It is commonly used to apply knowledge learned from one domain to another related but not identical domain. \cite{chen2021data, orbesarteaga2022augmentation}
\end{enumerate}




Knowledge-based engineering, a form of data augmentation, involves manipulating existing data features to generate new samples. While widely used, knowledge-based data augmentation poses both advantages and limitations.
\begin{enumerate}
    \item Limitations:
    
    Performance gain can be marginal when data is sufficient. In cases where the original dataset is sufficient, the incremental performance improvement achieved through knowledge-based engineering may be minimal\cite{eda}. Besides, knowledge-based engineering often involves generating synthetic samples by synonym replacement or structural adjustments \cite{eda, aeda, zhang2016characterlevel}, which will not change the label of the original dataset, so the imbalance in the dataset remains.
    \item Advantages:
    
    Amidst its limitations, feature engineering offers several advantages. Firstly, it is fast and simple. Knowledge-based engineering techniques are often straightforward to implement, requiring minimal computational resources and expertise. Secondly, performance gain is clear for small datasets. Conversely, in scenarios with limited data, knowledge-based engineering can lead to significant performance improvements by diversifying the training set.
\end{enumerate}

\subsection{Model-Based}
In this section, we categorize model-based data augmentation strategies into three distinct approaches: 1) Direct Training, which involves training a model exclusively on task-specific data for the purpose of data augmentation. 2) Pre-train and Fine-tune, which utilizes a model that has been pre-trained on a general dataset and subsequently fine-tuned with task-specific data to enhance data augmentation. 3) Using Foundation Models without Fine-tuning, which employs pre-trained models, prompted directly, to generate new data. Each method leverages different aspects of model training and adaptation to increase dataset diversity and improve model performance.

\subsubsection{Direct Training}
Before the widespread adoption of pre-trained models, we often develop a model that is trained exclusively on data specific to the task at hand for synthesizing new data. For example, if the task is image classification for detecting dogs, the model would only be trained on images of dogs. Once trained, this model can be used to generate new data samples that mimic the training data. This could involve techniques like generative adversarial networks (GANs, \citealp{goodfellow2014generative}) that can create entirely new images for augmentation \citep{antoniou2017data}. The key characteristic of this approach is that the augmentation model does not leverage any pre-existing models or datasets; it starts from scratch, learning exclusively from the task-specific dataset.

A typical example is when performing back-translation (\citealp{sennrich-etal-2016-improving}; \citealp{wieting-etal-2017-learning}; \citealp{mallinson-etal-2017-paraphrasing}), a technique that is particularly useful in the field of neural machine translation (NMT). Initially, an existing translation dataset is employed to train a neural translation model (NTM). Once trained, this NTM is used to translate the original dataset from the source language into one or more target languages. Subsequently, a second NTM, which might be the same or a different model trained in the reverse direction, translates these foreign language texts back into the source language. This process essentially generates additional, synthetic text data in the source language, which can be used to further train the translation model, thereby improving its accuracy and robustness through what is effectively a form of data augmentation.

Another example is when conducting pseudo-labeling (\citealp{lee2013pseudo}; \citealp{shi2018transductive}; \citealp{iscen2019label}; \citealp{arazo2020pseudo}), a semi-supervised learning technique used when there is a large amount of unlabeled data and a smaller set of labeled data. The process begins by training an initial model strictly on the available labeled data, which is then employed to make predictions on the unlabeled data. These predictions, despite not being verified by human annotation, are treated as true labels (hence the term "pseudo-labeling") and used to expand the training dataset. This augmented dataset, now containing both originally labeled and pseudo-labeled data, is used to retrain the model, potentially enhancing its performance due to the increased volume and variety of training data.


\citet{Fadaee_2017} generated synthetic data by targeting a word and replacing by a rare word generated by an LSTM model, which is trained on large amounts of monolingual data in both forward and backward directions. Generating rare words improves the diversity of datasets and leads to higher translation quality. Another implementation of RNN involves simply masking a word in a sentence, then generating the masked context with a bidirectional RNN language model \cite{kobayashi2018contextual}. This approach is independent of the NLP task, giving a general method for various domains. \citet{xu2016improved} proposed a data augmentation method by leveraging the directionality of relations through RNN for relation classification. The authors further compared their method with various model architectures and got the best performance. \citet{guo2019augmenting} used two approaches: one performs interpolation on words to mix up word embedding, and the other focuses on mixing up sentences using CNN or LSTM. They state that the interpolation strategies are a simple yet effective data augmentation method.

Different from the knowledge-based engineering approach to data augmentation, the current technique introduces the concept of models to improve the method of generating synthetic data.

\begin{enumerate}
    \item Advantages:    
    Trained models generate more diverse and realistic data than knowledge-based engineering methods, which can help improve the robustness and generalization of the main model. These models can be trained on specific domains or tasks, making them more adaptable to various datasets and requirements. As a result, they can perform complex transformations that basic methods cannot achieve. For instance, in image augmentation, a trained model can create entirely new images with different backgrounds, lighting conditions, and objects.
    \item Limitations:
    The main limitation of the current technique is its reliance on large amounts of labeled data for training, which is not always readily available. Additionally, training models requires significant computational resources and time, especially for large datasets or complex tasks, making it less efficient compared to simpler rule-based methods. Finally, these augmentation models can still overfit the training data, producing repetitive data patterns that may not accurately reflect real-world data distributions.
\end{enumerate}

\subsubsection{Pre-train then Fine-tune}
This section covers augmentation techniques under the pre-train then fine-tune paradigm. These are useful when labeled data is limited or transferring knowledge across tasks is beneficial. During pre-training, a model learns meaningful data representations on a large related dataset using unsupervised learning. Then, the pre-trained model is fine-tuned on a smaller labeled dataset for the target task, adapting its parameters to that specific task.

In the realm of NLP task, AUGNLG (\citealp{xu2021augnlg}) combine a self-trained neural retrieval model with a few-shot learned natural language understanding (NLU) model to generate MR-to-Text (meaning representation to text) data from open-domain texts, facilitating data augmentation in natural language processing (NLP) tasks. In the realm of synthetic data generation, various studies have demonstrated the efficacy of using diffusion models to enhance model performance, particularly under conditions where labeled data is scarce. \citet{doubinsky2023semantic} explore the potential of synthetic data in enhancing few-shot class-agnostic counting. They employ a dual conditioning approach using Stable Diffusion \citep{rombach2022high}, incorporating both a prompt and a density map to augment the training dataset for few-shot counting. Moreover, they enhance the diversity of synthesized images by implementing an exchange of captions between images.In the context of few-shot learning, leveraging synthetic data proves to be particularly advantageous. SeedSelect (\citealp{samuel2023generating}) observe a common failure of text-to-image models in generating rare concepts present in the training data. This issue can be mitigated by judiciously selecting generation seeds in the noise space, utilizing a small reference set of images. The incorporation of semantically appropriate generated images significantly enhances performance in few-shot recognition benchmarks. Task-specific augmentation further highlights the tailored application of these techniques. DiffTumor (\citealp{chen2024generalizable}) observe that early-stage tumors often exhibit similar imaging characteristics in computed tomography. To address this, the authors propose a multi-stage training pipeline to adapt the diffusion model, enabling the generation of realistic tumor images across various organs, based on arbitrary masks.

\begin{enumerate}
    \item Advantages:
    
    Pre-trained models leverage self-supervised learning to utilize knowledge acquired during the pre-training phase. Consequently, compared to direct training, pre-trained models don't require extensive data for fine-tuning to achieve similar or even superior performance. Especially when used for data augmentation, there is often a shortage of data. At such times, the advantage of pre-trained models comes into play. Moreover, the same pre-trained model can be fine-tuned with different model heads attached, using various datasets to accomplish different downstream tasks.
    \item Limitations:
    
    Pre-trained models are prone to overfitting on small amounts of data, leading to domain shift when used for data augmentation. Additionally, when fine-tuning pre-trained models, it's crucial to carefully adjust hyperparameters. While a particular set of settings might work for data augmentation on one dataset, it may not yield the same results on others \citep{kumar-etal-2020-data}.
    
\end{enumerate}

\subsubsection{Foundation Models without Fine-tuning}
As more data and advanced techniques are used for model pre-training, these pre-trained models showcase a greater range of possibilities. These models often exhibit excellent performance on downstream tasks without the need for additional fine-tuning. The emergence of these technologies has also provided us with different viewpoints when it comes to data augmentation.

Many of the current pre-trained language models (PLMs) have demonstrated their ability in commonsense reasoning within zero-shot scenario. Phi-series models (\citealp{gunasekar2023textbooks}; \citealp{li2023textbooks}; \citealp{abdin2024phi}) gather high-quality "textbook" data from the web, supplemented by synthetically generated data using GPT-3.5 \citep{achiam2023gpt}, to train their small-sized Transformer-based model. Evol-Instruct \citep{xu2023wizardlm} generates large amounts of instruction data with diverse levels of complexity using PLMs rather than relying on humans. \citet{auggpt} utilized ChatGPT \cite{chatgpt} to paraphrase samples from the training data and generate conceptually similar but semantically different samples. \citet{abdullin2024synthetic} generated synthetic dialogue dataset by allowing two LLM agents to engage in conversation. Moreover, these PLMs have also showed that they can learn through a smaller number of in-context examples, known as in-context learning \cite{gpt3-paper, LLM-survey, in-context-survey-dong}. Through in-context learning, it becomes easier for PLMs to generate high-quality synthetic data following predefined formats, thus reducing the complexity of post-processing \cite{z-icl}. For example, \citet{liu-etal-2022-wanli} and \citet{sahu-etal-2022-data} employed in-context learning to generate natural language inference and intent classification data, respectively. \citet{lee-etal-2022-personachatgen} utilized profile sentences to enable PLMs to generate profile sentences for different persona categories, thereby aiding in the generation of synthetic personalized dialogue datasets. \citet{wang-etal-2023-self-instruct} and \citet{honovich-etal-2023-unnatural} only collected a small amount of instruction-following data manually and then used in-context prompting to enable PLMs to generate large-scale synthetic instruction-following datasets. They found that this approach increased the diversity of the dataset.

Diffusion models have advanced the development of synthetic images for various applications, including fine-grained classification \citep{dunlap2023diversify} and semantic segmentation \citep{wu2024gptprompt}. In these approaches, a PLM like GPT generates image editing prompts, which are then used by a diffusion model to produce synthetic images that help train downstream models. CamDiff \citep{luo2023camdiff} focuses on augmenting camouflage object detection (COD) datasets with salient objects, thereby improving the robustness of COD models. In contrast, VIXEN \citep{black2024vixen} tackles the issue of limited training data and manipulation variety in Image Difference Captioning (IDC) datasets by using synthetically manipulated images from the recently developed InstructPix2Pix \citep{brooks2023instructpix2pix} dataset. These strategies not only demonstrate the versatility of diffusion models in various applications but also highlight a collective movement towards more dynamic and adaptable training datasets in machine learning research.

\begin{enumerate}
    \item Advantages: Using foundation models directly allows for quicker deployment because there's no need for an additional fine-tuning phase. This is especially beneficial in time-sensitive scenarios. Skipping the fine-tuning process also reduces computational costs and resource usage, which is significant when working with very large models. Additionally, foundation models trained on extensive and diverse datasets may already possess the necessary knowledge and patterns to generate high-quality synthetic data across a broad range of topics without further specialization.
    
    \item Limitations: The synthetic data generated by foundation models may not be as tailored to specific domain needs compared to data from fine-tuned models. This can lead to less accurate or less effective data for training downstream models. Foundation models applied directly might produce data with biases or inaccuracies that are not immediately apparent, as the data generation is not optimized for a specific task or domain. Controlling or influencing the nature of the generated data is also more challenging when using a foundation model directly, unlike fine-tuning the model on a particular dataset to reflect desired characteristics.
\end{enumerate}




\begin{table*}
    \small
    \centering
    \begin{tabular}{p{0.18\textwidth}p{0.37\textwidth}p{0.37\textwidth}}
        \toprule
        Objective & Concept & Examples \\
        \midrule
        Basic Quality & Basic quality aims to focus on quality of dataset. In the realm of NLP dataset, quality of datasets may include fluency, grammatical accuracy, format validation and so on. & \citet{DBLP:journals/corr/abs-1911-03324},\citet{kann2018sentencelevel}, \citet{lee-etal-2022-personachatgen}, \citet{lee-etal-2022-personachatgen}, \citet{gao2020paraphrase}, \citet{abdullin2024synthetic}\\
        \hline
        Label Consistency & Since discrepancies between generated data and their corresponding labels can undermine model performance and lead to incorrect inferences, it is crucial to address a series of post-processing to avoid label inconsistency of the synthetic data.& \citet{ChineaRios2017AdaptingNM}, \citet{anabytavor2019data},\citet{zhou2022flipda}, \citet{puri2020training}, \citet{Liu2021MulDAAM}, \citet{dall-e-for-detection}\\
        \hline
        Data Distribution & Data distribution focuses on maintaining distribution consistency, addressing domain shift, or avoiding similar data. & \citet{wang-etal-2023-self-instruct}, \citet{YU2024102253, gao2020paraphrase}, \citet{Yang_2020}, \citet{shakeri2020endtoend}, \citet{app13179766}, \citet{liu2022wanli}, \citet{wang-etal-2023-self-instruct}, \citet{dialogcc}, \citet{thakur2021augmented} \\
        \bottomrule
    \end{tabular}
    \caption{The purpose for doing post-processing and the related examples}
    \label{tab:post_processing}
\end{table*}

\section{Post-processing}\label{section:filtering}
After augmenting synthetic data, it is important to further filter or evaluate the dataset to ensure that the synthetic data is beneficial for the tasks and will improve the performence. The purpose of post-processing varies across tasks and situations. For instance, some post-processing efforts focus on filtering basic quality, such as the fluency and grammatical accuracy of sentences. Others may concentrate on obtaining intended data distribution, either to increase generalizability or to transfer the distribution to a specific domain. As shown in table \ref{tab:post_processing}, we address three critical purposes for doing post-processing: \textit{basic quality}, \textit{label consistency}, and \textit{data distribution}, which we will describe in detail in the subsections.

\subsection{Basic Quality}
Basic quality encompasses elements such as fluency, grammatical accuracy, format validation among others.

To assess the basic quality of dataset, various NLP metrics are commonly used.  \cite{DBLP:journals/corr/abs-1911-03324} ensures the integrity and relevance of the content by setting high-quality thresholds based on summarization-specific metrics like oracle scores and ROUGE-2 recall. \citet{kann2018sentencelevel} introduces a syntactic log-odds ratio (SLOR) to evaluate fluency and SLOR and is used in evaluating the synthetic dataset in \citet{feng2020genaug}. Regular expressions were utilized to ensure the correct format of the output, and the output data was compared with the in-context examples to filter out any instances of repeated data \citep{lee-etal-2022-personachatgen}.

 When generating synthetic persona-based dialogue, \citet{lee-etal-2022-personachatgen} ensures persona consistency through a fine-tuned RoBERTa-based NLI model. Also, pre-trained vision-language models are often employed to validate synthetic multi-modal data. \citet{gao2020paraphrase} discusses a Paraphrase Augmented Response Generation (PARG) framework that enhances dialogue generation by training a paraphrase and response generation model together. The data filtering technique  focuses on selecting high-quality paraphrase pairs based on their semantic similarity and surface form diversity.
 In another paper by \citet{abdullin2024synthetic}, the authors employ a prompt to request GPT-4 to mimic human evaluation methods evaluating the readability of generated text.

\subsection{Label Consistency}
When generating synthetic data with labels, there is a possibility of discrepancies between the data and its labels. To avoid these inconsistencies, certain post-processing steps will be implemented following data augmentation to maintain label accuracy throughout the dataset.

\citet{ChineaRios2017AdaptingNM} presents a data filtering technique for adapting neural machine translation systems, utilizing vector space representations of sentences. It employs a dynamic threshold for cosine similarity to select synthetic sentences that are closely aligned with the centroid of a test set, ensuring label consistency. 

The study by \citet{anabytavor2019data} involves class labeling, where the authors train a classifier using existing labeled data. They then use this classifier to filter generated text, which is produced using GPT. \citet{dall-e-for-detection} utilized CLIP to ensure that interest classes are not present in the generated context description images (CDI). \citet{Liu2021MulDAAM} presents a data filtering approach in a multilingual data augmentation framework for named entity recognition (NER), focusing on enhancing label consistency. \citet{puri2020training}, they employ a roundtrip filtration method. This involves using a pre-trained QA model to infer answers for the generated triplets (QPA). The consistency between the inferred answers and the generated answers is then assessed. If they are consistent, the generated triplet is retained. \citet{zhou2022flipda} trained a classifier to assess whether the label is consistence with the augmented data label.

\subsection{Data Distribution}

When utilizing the LLMs' in-context learning capability to generate synthetic data, there is often a risk of encountering copy-paste behavior or generating data that closely resembles the in-context examples. Maintaining the distribution consistency or focusing on domain shift also benefits from some post-processing. Focusing on data distribution when filtering ensures the effectiveness of data augmentation.

\citet{wang-etal-2023-self-instruct} employed Rouge-L to compare the generated instruction-following data with the instruction-following data in the task pool. They filtered out data with excessively high similarity to ensure the diversity of synthetic data within the task pool. \citet{YU2024102253, gao2020paraphrase} filter and evaluate synthetic data by NLP metrics, utilizing BLEU scores to assess semantic relevance and diversity scores to evaluate surface form variation. The filtering method \citet{Yang_2020} proposed, named G-DAUGc-Influence, removes detrimental synthetic data by analyzing their influence on validation loss. \citet{shakeri2020endtoend} address the task of QA and critique the efficiency of previous filtering methods, which predominantly relied on pre-trained QA models for selection. To enhance efficiency, they propose a novel filtering strategy that utilizes a Language Model score, based on the relevance between the answer, context, and question, as a metric to filter and select generated data. \citet{app13179766} directly addresses issues of class imbalance by enriching the dataset with diverse and novel synthetic reviews. To ensure the dataset to be diverse, they filtered out data by similarity score.  

In \citet{liu2022wanli} work, they filtered the generated examples to keep the most ambiguous ones based on the model. \citet{wang-etal-2023-self-instruct} employed Rouge-L to compare the generated instruction-following data with the instruction-following data in the task pool. They filtered out data with excessively high similarity to ensure the diversity of synthetic data within the task pool. \citet{dialogcc} utilized CLIP to compute the similarity between dialogue turns and images when generating synthetic visual dialogue, ensuring text-image alignment. The study by \citet{thakur2021augmented} introduces a sampling strategy that encompasses multiple methods, with BM25 Sampling being identified as the most efficient. Utilizing ElasticSearch, this approach involves extracting the top k most similar sentences for each given sentence as part of its filtering strategy.

\section{Future Work}
This section examines the shift in data augmentation from focusing on quantity to emphasizing quality to enhance machine learning model performance. We explore how augmented data can enrich dataset diversity, assess biases, and address distribution shifts. Additionally, we discuss adapting data distributions with synthetic data to better suit specific tasks, and outline the need for developing standardized benchmarks to evaluate these methods. The discussion also touches on the integration of diverse data types in multi-modal data augmentation, highlighting new challenges and opportunities for advancing machine learning models.

\subsection{From Quantity to Quality}

In the past, data augmentation methods have primarily focused on increasing the quantity of data to enhance model performance, particularly when datasets are small or lack diversity. This approach can significantly improve a model’s generalization ability. However, the benefits of adding more data are not infinite. As the volume of data reaches a certain threshold, the incremental gains in model performance begin to diminish. This phenomenon is known as "diminishing returns." Given these constraints—rising costs and increased training time—the emerging trend is toward enabling models to learn effectively from smaller but high quality datasets, achieving performance levels comparable to those obtained from larger but low quality datasets. Therefore, the suggested approach lies in focus more on quality rather than quantity\cite{schimanski2024faithful}. There are some approaches we suggest: 1) Enhancing the quality of synthetic data: reducing the generation of invalid data or improving post-processing techniques. 2) Expanding the coverage of knowledge dimensions in synthetic datasets: enabling models to learn a wide range of knowledge from a smaller amount of data to enhance the model's generalization ability.

\subsection{The Evaluation of Augmented Data}

Creating a standard benchmark for evaluating data augmentation techniques—focusing on their quality, diversity, and relevance—is a key but complex challenge in advancing machine learning. Currently, the evaluation of these methods often relies on different datasets and metrics, without a consistent approach. Although benchmarks like CIFAR-10 and ImageNet provide standardized datasets for assessing techniques in certain areas, they may not fully capture the variety of challenges found in practical scenarios. Furthermore, evaluating the quality and relevance of data augmentation is subjective, meaning opinions on what counts as "good" augmentation can vary widely depending on the task and dataset. There's a need for new metrics that can measure both the tangible and intangible aspects of data augmentation to truly determine its effectiveness. Additionally, making sure this standard can grow and adapt to accommodate new techniques and varied application areas is another big challenge. Despite these difficulties, creating a strong benchmark is crucial as it could greatly help in developing more effective and flexible augmentation methods.


\subsection{Multi-modal data augmentation}
Currently, there are relatively few studies that focus on multi-modal data augmentation, even though this area holds significant potential for enhancing model performance in complex tasks. Multi-modal data, which combines different types of data like text and images, presents unique challenges and opportunities for augmentation. By developing new methods in this field, researchers can better address the intricacies of integrating diverse data types, leading to more sophisticated and capable models.Take vision-text tasks as example. MixGen \cite{hao2023mixgen} is the state-of-the art augmentation method for vision language modalities and generates new image-text pairs by linear interpolating between two images and concatenating two texts. LeMDA \cite{liu2023learning} is an method that learns to jointly augment multi-modal data in feature space. 

All the prior works on multi-modal data augmentation assume a pre-existing alignment between the modalities they augment. This assumption overlooks a critical aspect: the instances where the modalities are misaligned or where the relationship between them is not straightforward. Addressing this oversight could unlock further potential in multi-modal applications by developing augmentation techniques that also consider and enhance the non-aligned portions of the data. This gap signifies an opportunity for novel research directions that could lead to more robust models capable of handling diverse and complex multi-modal scenarios.

\section{Conclusion}

The paper explores contemporary synthetic data techniques from the perspectives of augmented objectives, different technological eras, and the purposes of post-processing. Additionally, we identify three future directions: 1) \textit{From Quantity to Quality}, 2) \textit{The Evaluation of Augmented Data}, 3) \textit{Multi-modal data augmentation}. We hope that these insights will help the research community in future studies on synthetic data.

\section*{Acknowledgements}
This work was financially supported by the National Science and Technology Council (NSTC) in Taiwan, under Grants 112-2223-E-002-012-MY5 and 111-2222-E-002-013-MY3, and from the Featured Area Research Center Program within the framework of the Higher Education Sprout Project by the Ministry of Education (113L900901/113L900902/113L900903).



\bibliography{anthology,custom}
\appendix



\end{document}